\documentclass[format=sigconf, authorversion=true]{acmart}
\usepackage[font=small]{caption}

\begin{document}
\title{Large-Scale Plant Classification with Deep Neural Networks}
\acmConference[ACM CF'17]{ACM International Conference on Computing Frontiers}{May 15-17, 2017}{Siena, Italy}
\author{Ignacio Heredia}

\orcid{0000-0001-6317-7100}
\affiliation{%
\vspace*{10pt}
\institution{\textsl{\normalsize Instituto de Fisica de Cantabria (CSIC-UC)}}
\streetaddress{\textsl{\normalsize E-39005, Santander, Spain}}
}

\authornote{Electronic address: \texttt{iheredia@ifca.unican.es}}


\copyrightyear{2017}
\acmYear{2017}
\setcopyright{acmlicensed}
\acmConference{ACM CF'17}{May 15-17, 2017}{Siena, Italy}\acmPrice{15.00}\acmDOI{10.1145/3075564.3075590}
\acmISBN{978-1-4503-4487-6/17/05}

\begin{abstract}
This paper discusses the potential of applying deep learning techniques
for plant classification and its usage for citizen science in large-scale
biodiversity monitoring. We show that plant classification using near
state-of-the-art convolutional network architectures like ResNet50
achieves significant improvements in accuracy compared to the most
widespread plant classification application in test sets composed
of thousands of different species labels. We find that the predictions
can be confidently used as a baseline classification in citizen science
communities like iNaturalist (or its Spanish fork, Natusfera) which
in turn can share their data with biodiversity portals like GBIF.
\end{abstract}
\maketitle
\renewcommand{\figurename}{FIG.}

\section{Introduction}

The deep learning revolution has brought significant advances in a
number of fields \cite{LeCun2015}, primarily linked to image and
speech recognition. The standardization of image classification tasks
like the ImageNet Large Scale Visual Recognition Challenge \cite{ILSVRC15}
has resulted in a reliable way to compare top performing architectures.
Since the AlexNet architecture \cite{AlexNet}, the first efficient
implementation of convolutional neural networks using GPUs, the error
in these competitions has reached superhuman performance \cite{He2015_superhuman}. 

Despite this recent success in general image recognition, the work
in the biodiversity community relies heavily on hand labeled image
data assigned by a (relatively) small community of experts and does
not exploit these recent advances. This might be an impediment to
open the community to non expert users who, armed with modern technologies
handily embedded in a smartphone, can push biodiversity monitoring
to the next level. The use of deep learning for plant classification
is not novel \cite{DeepPlant,dyrmann2016plant} but has mainly focused
in leaves and has been restricted to a limited amount of species,
therefore making it of limited use for large-scale biodiversity monitoring
purposes. This same specificity issue applies to some standardized
plant datasets \cite{Nilsback08/Oxford102} which are very helpful
to evaluate the network performances but who are limited in variety
of species or in the diversity of the images (focusing mainly in flowers
or leaves). The PlantNet tool \cite{Bonnet2015PlantNet,Joly2014_plantnet2},
based on distant versions of the IKONA algorithms, pioneered in creating
an open access tool to automate the task of recognizing a wide variety
of species. However it does not reach the performance of expert botanists.
Applying the recent advances in convolutional neural networks could
have a positive impact in closing this performance gap. This could
be a large step towards building a reliable and general large-scale
plant recognition app that spreads the use of citizen science for
biodiversity monitoring. 

\section{The dataset}

As training dataset we use the great collection of images which are
available in PlantNet under a Creative-Common Attribution-ShareAlike
2.0 license. It consists of around 250K images belonging to more than
6K plant species of Western Europe. These species are distributed
in 1500 genera and 200 families. Each image has been labeled by experts
and comes with a tag which specifies the focus of the image, like
'habit', 'flower', 'leaf', 'bark', etc. Most images have resolutions
ranging from 200K to 600K pixels and aspect ratios ranging from 0.5
to 2. The dataset is highly unbalanced because most labels contain
very few images. 

We train on the whole dataset (without making validation or test splits)
as we intend to build a classifier trained on the same dataset as
the PlantNet tool so that their performances can be fairly compared.
Also we believe that testing the classification performance on a subset
of PlantNet is not an accurate measure of the performance of the net
on real-world data as all the images in the dataset are highly correlated
(many photos inside a specie share author and are often taken from
the same plant with slightly different angles). Therefore at test
time we will use three external datasets to confidently measure the
performance of our net. 

\section{The model }

As plant classification is not very different from general object
classification, we expect that top performing architectures in the
ImageNet Large Scale Visual Recognition Challenge (ILSVRC) would perform
well in this task. Therefore we use as convolutional neural network
architecture the ResNet model \cite{ResNet} who won the ILSVRC'15.
This architecture consists of a stack of similar (so-called residual)
blocks, each block being in turn a stack of convolutional layers.
The innovation is that the output of a block is also connected with
its own input through an identity mapping path. This alleviates the
vanishing gradient problem, improving the gradient backward flow in
the network and allowing to train much deeper networks. We choose
our model to have 50 convolutional layers (aka. ResNet50). 

As deep learning framework we use the Lasagne \cite{lasagne} module
built on top of Theano \cite{bergstra+al:2010-scipy,Bastien-Theano-2012}.
We initialize the weights of the model with the pretrained weights
on the ImageNet dataset provided in the Lasagne Model Zoo. We train
the model for 100 epochs on a GTX 1080 for 6 days using Adam \cite{Adam}
as learning rule. During training we apply standard data augmentation
(as sheer, translation, mirror, etc) so that the network never sees
the same image. We do not apply rotation or upside down mirroring
to the images tagged as 'habit', as it does not make much sense to
have a tree or a landscape upside down. After applying the transformations
we downscale the image to the ResNet standard 224$\times$224 input
size. \footnote{Code is available at \texttt{github.com/IgnacioHeredia/plant\_classification}}

\section{Experiments and Discussion}

Our goal is to achieve a performance that we consider to be useful
as baseline classification (ie. around 50\% accuracy). However it
is difficult to assess how our net performance compares to other existing
algorithms for plant classification as the main competition for plant
classification, PlantCLEF \cite{PlantClef2016}, uses datasets composed
of images uploaded to PlantNet by users who might already be present
in our training set. Therefore we have composed three test datasets
with external photos.

To put the Resnet accuracy values into perspective, we will compare
them with the performance of the PlantNet tool on these same three
datasets. In the PlantNet tool you can upload an url, or an image
from your local disk, along with a tag suggestion and it returns a
list of suggested species. When assessing its performance we report
the best predictions across all tags (ie. we suppose the user selects
optimally the tag). 

Finally for the ResNet50 evaluation we use random ten crop testing
with smaller data augmentation parameters than those used during training.

\subsection{The datasets}

\subsubsection{Google Search Image}

For this dataset we select the 3680 labels (around 60\% of all labels)
with more than 12 images in our training dataset. For each one of
these labels we automatically retrieve the 10 first images returned
by the Google Image Search engine. As this is done in an automated
fashion some minor mislabeled or corrupt examples might appear in
the dataset. By choosing only the most popular labels and retrieving
the top results, we expect to minimize the presence of mislabeled
images.

\subsubsection{Portuguese Flora}

The Portuguese flora dataset \cite{Flora-on.pt} consists in 23K images
belonging to 2K species. To compose our test dataset we select the
15K images belonging to one of the 1300 species which are also present
in our training dataset.

\subsubsection{iNaturalist}

iNaturalist is a website were the user can upload their observations,
that can have one or several images, and get help from the community
to have them correctly labeled. For composing our dataset we select
only the observations with research quality grade (ie. a consensus
has been reached in the community on the species or genus label).
There are around 600K such plant observations belonging to several
ranks like species (97\% of the total), genus, variety, subspecies,
hybrid, etc. Selecting the observations tagged as (pure) species we
end up with 900K images belonging to 20K plant species. From this
set of images we only select the ones belonging to any of our 6K training
species and we end up with as test set composed of 300K images belonging
to 3K different species.

In a later stage we will see how the prediction accuracy improves
with observations containing 2 images or more. For that we end up
with a test set of 60K observations containing between 2 and 33 images
belonging to 2600 species present in our training dataset.

\subsection{Results and Discussion}

\begin{figure}
\noindent\makebox{\includegraphics[scale=0.13]{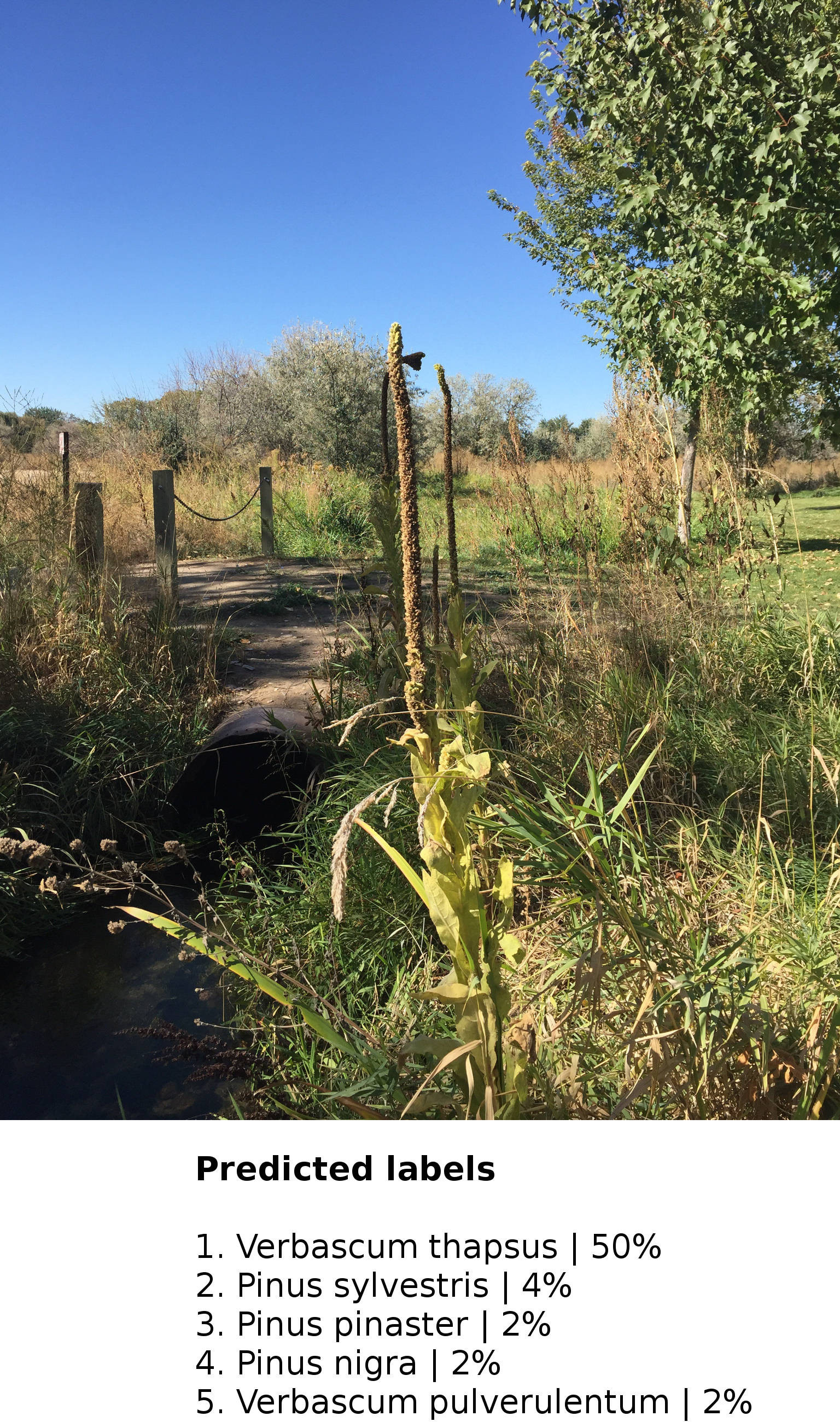}}

\caption{\textbf{\label{image-demo}}Example of non-trivial image classification
with the ResNet50. Here the true label is Verbascum Thapsus which
is also the first predicted label.}
\end{figure}

\begin{table}
\begin{center}
\begin{tabular}{ c c c c c } 
\hline
\noalign{\smallskip}
Datasets & \multicolumn{4}{ c }{ Accuracy \%} \\ 
\noalign{\smallskip}
         & \multicolumn{2}{ c }{ResNet50 (ours)} & \multicolumn{2}{ c }{PlantNet (usual)} \\ 
\noalign{\smallskip}
         & Top1 & Top5                         & Top1 & Top5                    \\   
\hline
\noalign{\smallskip}
Google Search  & 40 & 63 & 18 & 37 \\ 
\noalign{\smallskip}
Portuguese Flora     & 29 & 47 & 15 & 29 \\ 
\noalign{\smallskip}
iNaturalist          & 33 & 49 & 18 & 30 \\ 
\hline
\end{tabular}
\end{center}
\caption{\label{tab:Accuracy results}Accuracy results of the two algorithms
for all three test datasets. }
\end{table}

The ResNet50 returns a list of probabilities that each label is the
correct label as shown in Fig \ref{image-demo}. The top1 accuracy
measures how often the correct label is the highest probability label,
while the top5 accuracy measures how often the correct label is among
the five labels with highest probability. Table \ref{tab:Accuracy results}
shows the top1 and top5 accuracy results for all three datasets. We
can notice that the Resnet50 achieves $\times$2 and $\times$1.7
improvements for top1 and top5 accuracies consistently across datasets
compared with the PlantNet tool. The overall accuracy is approximately
constant although slightly higher in the Google dataset probably due
to higher image quality.

\begin{figure}
\noindent\makebox{\includegraphics[scale=0.062]{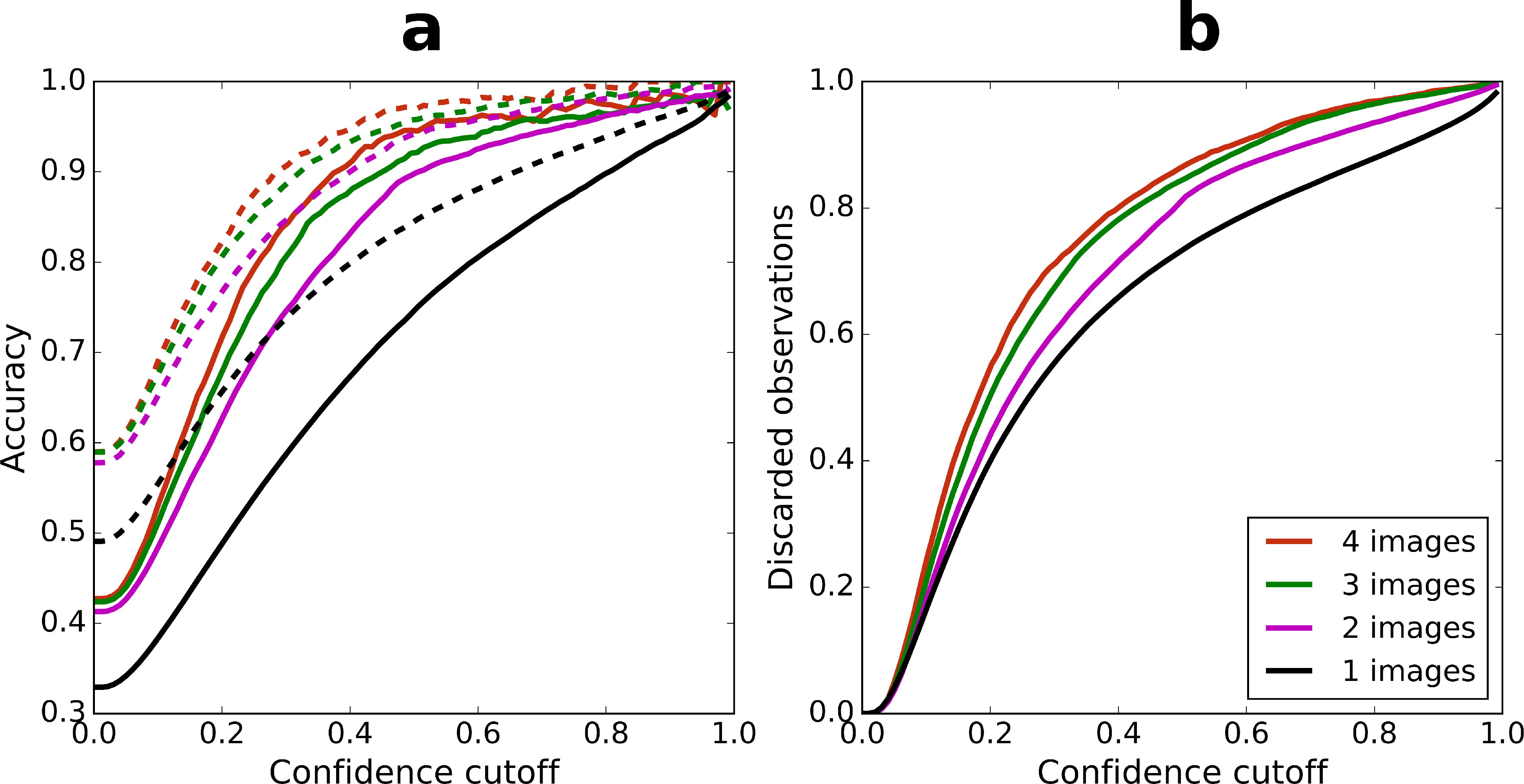}}

\caption{\textbf{\label{multiobservations}}Detailed results for the iNaturalist
dataset for observations containing from 1 to 4 images.\textbf{ a)}
Top1 (solid line) and Top5 (dashed line) accuracy as a function of
the probability of the first predicted label. \textbf{b)} Proportion
of observations that have to be discarded because they do not meet
the desired confidence.}
\end{figure}

Although the accuracy results are better than those obtained with
the PlantNet tool, they are far from being reliable enough to be systematically
used to predict tags for all observations. One way to improve this
is to only return an identification if the net is confident enough
about its prediction. The Fig \ref{multiobservations}a shows how
this accuracy improves when we only trust predictions who have a top1
probability above a certain cutoff. For example if we set the cutoff
at 30\% the top1 and top5 accuracies increase to 59\% and 74\% respectively.
The value of the cutoff should be a trade-off between how confident
we want to be and how many observations we are willing to discard.
In Fig \ref{multiobservations}b we show the proportion of observations
that had to be discarded because they did not meet the desired cutoff
probability. In the case of setting the cutoff to 30\%, we are discarding
55\% of the observations.

But increasing the confidence cutoff is not the only way to improve
the accuracy. Although observations with a single image are the majority
(91\% of the total) in the iNaturalist dataset, there are also observations
with 2 (6\%), 3 (2\%), 4 (1\%) and more images. If we use jointly
those images to identify average the predictions of the observed specie
we achieve much higher accuracies than with only one image due to
the lower influence of random noise. For example if we examine again
with the cutoff to 30\%, we now have top1 accuracies of 75\%, 80\%
and 84\% for observations with 2, 3 and 4 images respectively. However
the proportion of discarded images also increases compared to the
1 image case, reaching now 60\%, 67\% and 71\%. 

Although one might argue that those multi image observations are a
very small portion of the dataset (and therefore the improvement in
overall accuracy marginal), it is important to notice the increasing
trend of uploads of multi images observations in the recent times.
For example in the last three months of 2016, the 1 image observations
were merely 60\% of the total whereas the 2, 3 and 4 image observations
went up to 23\%, 11\% and 4\% respectively. 

\begin{figure*}
\noindent\makebox[\textwidth]{\includegraphics[scale=1.]{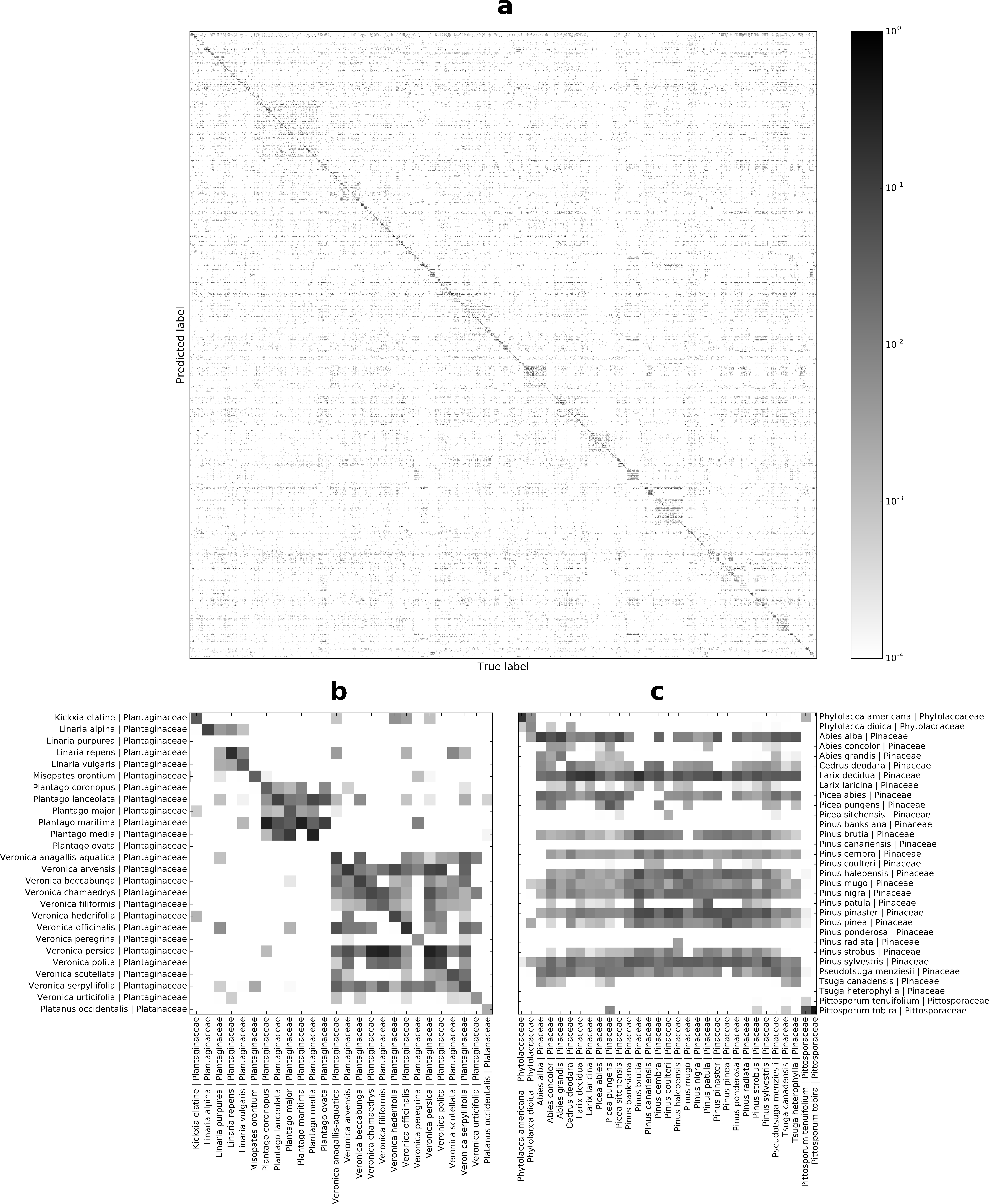}}

\caption{\textbf{\label{mixing}a)} Confusion matrix for the iNaturalist dataset
of observations with 1 image. We zoom \textbf{b)} in a region where
different genera as confused separately inside the same family and
\textbf{c)} in a region where all the genera are confused inside the
same family. We weight the counts in the matrix with the probability
of the prediction. All columns have been normalized to 1. We plot
only the 1.5K labels with more observations so that the matrix appears
more dense.}
\end{figure*}

In Fig \ref{mixing}a we show the confusion matrix of the iNaturalist
test dataset predictions for observations with one image. We have
ordered the species label in blocks of families and blocks of genera
to unravel the inherent block structure of the matrix. As we can see,
along the diagonal, which is densely populated as expected, there
are blocks of different sizes which denote groups inside which confusion
is frequent. Fig \ref{mixing}b zooms into one such a group where
we can see that inside the family of Plantaginaceae, the species belonging
to the genera Linaria, Plantago and Veronica are often confused with
other species within the same genera. Another example of a typical
block could be Fig \ref{mixing}c where we can see that the species
belonging to the family are often confused with other species of the
family irrespectively of their genus. Those three figures show that
even when the net's confidence is high but the prediction is wrong,
we will likely be able to extract useful information, either about
the correct genus or either about the correct family. This can be
valuable information for the users or experts to narrow the search
of the correct specie.

Finally, with regard to a future deployment in the iNaturalist ecosystem,
we have to note that those accuracy results are restricted to the
species present in our training dataset who merely represent 30\%
of all plant species present in iNaturalist. This is due to the fact
that we trained with just Western Europe species from PlantNet while
iNaturalist receives observations from all around the world. This
could be solved in future work by retraining the net with both the
iNaturalist and all the PlantNet images (including South America,
Indian Ocean and North Africa).

\section{Conclusion}

In this work we have built a large-scale plant classification algorithm
based on the ResNet convolutional neural network architecture. We
have evaluated the classification performance of our net on the observations
of iNaturalist and obtained that we were able to classify almost half
of these observations, who lied above a 30\% predictive cutoff, with
a top1 and top5 accuracies of 59\% and 74\% respectively. We have
then demonstrated that the user ability to upload several images per
observation (preferably of different plant parts or from different
angles) critically improved the final accuracy. Finally we obtained
that even when the prediction was wrong it was very likely that we
could obtain some information about the true genus or family, so that
it could be used by experts or users to narrow their search of the
correct label.

In addition we have seen that trained with the same image dataset,
the ResNet architecture outperforms the most widespread online public
plant classification algorithm by around a factor of 2 in top1 and
top5 accuracies. Besides our model does not require to enter a suggested
image tag along with the observation.

With all this information in hand we think that large-scale biodiversity
projects like the Global Biodiversity Information Facility (GBIF)
\cite{GBIF} or LifeWatch \cite{Lifewatch}, the European research
infrastructure on biodiversity, could very well benefit from this
new techniques to build a fast and reliable method to automatically
monitor biodiversity. This tool can definitely open the field to active
contributions of non expert users including citizen scientists.

For future work there are several ways to explore how to achieve an
increase in accuracy. The most obvious way is to increase the training
dataset size. It should be noted that iNaturalist contains even more
images than PlantNet so when training a net for deployment one should
combine both datasets to increase the predictions accuracy. Here we
trained with just the PlantNet dataset so that the comparison of performance
with the PlantNet tool would be fair. The other way is to implement
architectural modifications to the net that lead to a better generalization
error. Along this line two promising variants of the Resnet architecture
have recently appeared. The first one is the Stochastic Depth Network
\cite{Stochastic_depth_nets} in which we remove randomly some residual
blocks during training, allowing the net to be more robust for generalization
and to train deeper networks. The second more promising variant is
the DenseNet \cite{Densenets} in which the skip identity connections
are now connecting the residual blocks at all scales. Lastly it is
worth mentioning that the Resnet50 is a quite space-consuming architecture
so if we were to implement plant recognition app in embedded devices,
so that the user could identify without connecting to the Net, one
could use some recent modifications of shallower architectures that
offer almost as good performance with much less memory consumption
\cite{Squeezenets}. 

\section*{Acknowledgements}

I want to thank Jesus Marco de Lucas and Fernando Aguilar for their
helpful comments and supervision. I also wanted to thank PlantNet
and Flora-on for their image's open access policy, and the EGI-Engage
Lifewatch Competence Centre for their support. The author is funded
with the EU Youth Guarantee Initiative (Ministerio de Economia, Industria
y Competitividad, Secretaria de Estado de Investigacion, Desarollo
e Innovacion, through the Universidad de Cantabria).

\bibliographystyle{unsrt}
\bibliography{references}

\begin{thebibliography}{10}

\bibitem{LeCun2015}
Yann LeCun, Yoshua Bengio, and Geoffrey Hinton.
\newblock Deep learning.
\newblock {\em Nature}, 521(7553):436--444, may 2015.

\bibitem{ILSVRC15}
Olga Russakovsky et~al.
\newblock {ImageNet Large Scale Visual Recognition Challenge}.
\newblock {\em International Journal of Computer Vision (IJCV)},
  115(3):211--252, 2015.

\bibitem{AlexNet}
Alex Krizhevsky, Ilya Sutskever, and Geoffrey~E Hinton.
\newblock Imagenet classification with deep convolutional neural networks.
\newblock In F.~Pereira, C.~J.~C. Burges, L.~Bottou, and K.~Q. Weinberger,
  editors, {\em Advances in Neural Information Processing Systems 25}, pages
  1097--1105. Curran Associates, Inc., 2012.

\bibitem{He2015_superhuman}
Kaiming He, Xiangyu Zhang, Shaoqing Ren, and Jian Sun.
\newblock Delving deep into rectifiers: Surpassing human-level performance on
  imagenet classification, 2015.

\bibitem{DeepPlant}
Sue~Han Lee, Chee~Seng Chan, Paul Wilkin, and Paolo Remagnino.
\newblock Deep-plant: Plant identification with convolutional neural networks,
  2015.

\bibitem{dyrmann2016plant}
Mads Dyrmann, Henrik Karstoft, and Henrik~Skov Midtiby.
\newblock Plant species classification using deep convolutional neural network.
\newblock {\em Biosystems Engineering}, 151:72--80, 2016.

\bibitem{Nilsback08/Oxford102}
M-E. Nilsback and A.~Zisserman.
\newblock Automated flower classification over a large number of classes.
\newblock In {\em Proceedings of the Indian Conference on Computer Vision,
  Graphics and Image Processing}, Dec 2008.

\bibitem{Bonnet2015PlantNet}
Pierre Bonnet et~al.
\newblock Plant identification: man vs. machine.
\newblock {\em Multimedia Tools and Applications}, 75(3):1647--1665, jun 2015.

\bibitem{Joly2014_plantnet2}
Alexis Joly et~al.
\newblock Interactive plant identification based on social image data.
\newblock {\em Ecological Informatics}, 23:22--34, sep 2014.

\bibitem{ResNet}
Kaiming He, Xiangyu Zhang, Shaoqing Ren, and Jian Sun.
\newblock Deep residual learning for image recognition, 2015.

\bibitem{lasagne}
Sander Dieleman et~al.
\newblock Lasagne: First release., August 2015.

\bibitem{bergstra+al:2010-scipy}
James Bergstra et~al.
\newblock Theano: a {CPU} and {GPU} math expression compiler.
\newblock In {\em Proceedings of the Python for Scientific Computing Conference
  ({SciPy})}, June 2010.
\newblock Oral Presentation.

\bibitem{Bastien-Theano-2012}
Fr{\'{e}}d{\'{e}}ric Bastien et~al.
\newblock Theano: new features and speed improvements.
\newblock Deep Learning and Unsupervised Feature Learning NIPS 2012 Workshop,
  2012.

\bibitem{Adam}
Diederik Kingma and Jimmy Ba.
\newblock Adam: A method for stochastic optimization, 2014.

\bibitem{PlantClef2016}
Alexis Joly et~al.
\newblock {LifeCLEF} 2016: Multimedia life species identification challenges.
\newblock In {\em Lecture Notes in Computer Science}, pages 286--310. Springer
  Nature, 2016.

\bibitem{Flora-on.pt}
Sociedade~Portuguesa de~Bot{\^{a}}nica.
\newblock Flora-on: Flora de portugal interactiva, 2014.
\newblock \url{http://www.flora-on.pt/}.

\bibitem{GBIF}
Global Biodiversity Information~Facility (GBIF).
\newblock \url{http://www.gbif.org/}.

\bibitem{Lifewatch}
Lifewatch: e-science and technology european infrastructure for biodiversity
  and ecosystem research.
\newblock \url{http://www.lifewatch.eu}.

\bibitem{Stochastic_depth_nets}
Gao Huang, Yu~Sun, Zhuang Liu, Daniel Sedra, and Kilian Weinberger.
\newblock Deep networks with stochastic depth, 2016.

\bibitem{Densenets}
Gao Huang, Zhuang Liu, Kilian~Q. Weinberger, and Laurens van~der Maaten.
\newblock Densely connected convolutional networks, 2016.

\bibitem{Squeezenets}
Forrest~N. Iandola, Song Han, Matthew~W. Moskewicz, Khalid Ashraf, William~J.
  Dally, and Kurt Keutzer.
\newblock Squeezenet: Alexnet-level accuracy with 50x fewer parameters and
  <0.5mb model size, 2016.

\end{thebibliography}


\end{document}